\documentclass[letterpaper]{article} 
\usepackage[preprint]{aaai2027}  
\usepackage[hyphens]{url}  
\usepackage{graphicx} 
\usepackage{natbib}  
\usepackage{caption} 
\usepackage{amsmath}
\usepackage{amssymb}
\usepackage{booktabs}
\usepackage{multirow}
\usepackage{pifont}

\title{ReMemNav: Memory-Based Decision Correction and Target Verification for Zero-Shot Object Navigation}

\author{
    Feng Wu$^{1,*}$, Wei Zuo$^{1,*}$, Wenliang Yang$^{1}$, and Jun Xiao$^{1}$, Yang Liu$^{2,\dagger}$, Xinhua Zeng$^{1,\dagger}$
}
\affiliations{
    $^{1}$Fudan University\\
    $^{2}$Tongji University\\
    $^{*}$These authors contributed equally to this work.\\
    $^{\dagger}$Corresponding Author\\
    \texttt{fengwu25, wzuo25, wlyang25, jxiao23}@m.fudan.edu.cn;\\
    \texttt{yang\_liu@ieee.org};\\
    \texttt{zengxh@fudan.edu.cn}
}

\begin{document}

\maketitle

\begin{abstract}
Zero-shot object navigation requires agents to locate unseen
targets in unfamiliar environments without prior maps or
task-specific training. Despite the commonsense reasoning
ability of vision-language models (VLMs), existing mapless
navigators often suffer from repeated exploration and premature
stopping caused by limited historical context and false-positive
target predictions. We propose ReMemNav, a training-free
framework that combines lightweight semantic grounding,
memory-based decision correction, and target verification.
RAM-derived semantic priors guide panoramic direction
selection, while a geometry-triggered correction mechanism
retrieves historical scene descriptions when the selected
direction points toward previously visited regions. When a
target is predicted to be visible, ReMemNav verifies the
corresponding single-view observation before the final
approach. Depth-based action sampling then converts the
high-level decision into collision-free motion. Experiments on
HM3D and MP3D show that ReMemNav achieves higher success rates
and path efficiency than existing training-free zero-shot baselines.
Specifically, ReMemNav achieves absolute SR/SPL gains of
1.7\%/5.3\% on HM3D v0.1, 18.2\%/11.1\% on HM3D v0.2,
and 8.7\%/6.9\% on MP3D.
\end{abstract}

\section{Introduction}

Efficient autonomous navigation is fundamental for domestic robots to assist with various tasks in unfamiliar and complex environments \cite{1}. As one of the most challenging tasks in this field, zero-shot object navigation requires an agent to actively locate and navigate to unseen target objects in completely unknown environments without prior maps or task-specific fine-tuning \cite{2}.

Existing zero-shot navigation methods generally follow two paradigms. The first relies on pretrained vision-language models (VLMs) for cross-modal feature alignment and guides navigation using image-text similarity \cite{2,khandelwal2022simple,gadre2023cows}. The second constructs spatial or semantic maps during exploration and performs navigation based on the resulting environmental representations \cite{7,huang2023visual,zhang2024imagine,6,yokoyama2024vlfm,kuang2024openfmnav}. However, feature-alignment methods often lack sufficient temporal and spatial context, whereas map-based approaches depend on accurate environmental representations and may introduce considerable construction and maintenance overhead. Recent advances in VLMs offer an alternative direction by enabling agents to directly make mapless navigation decisions using multimodal observations, world knowledge, and commonsense reasoning \cite{goetting2024end,5}.

\begin{figure}[t]
	\centering
	\includegraphics[width=0.95\columnwidth]{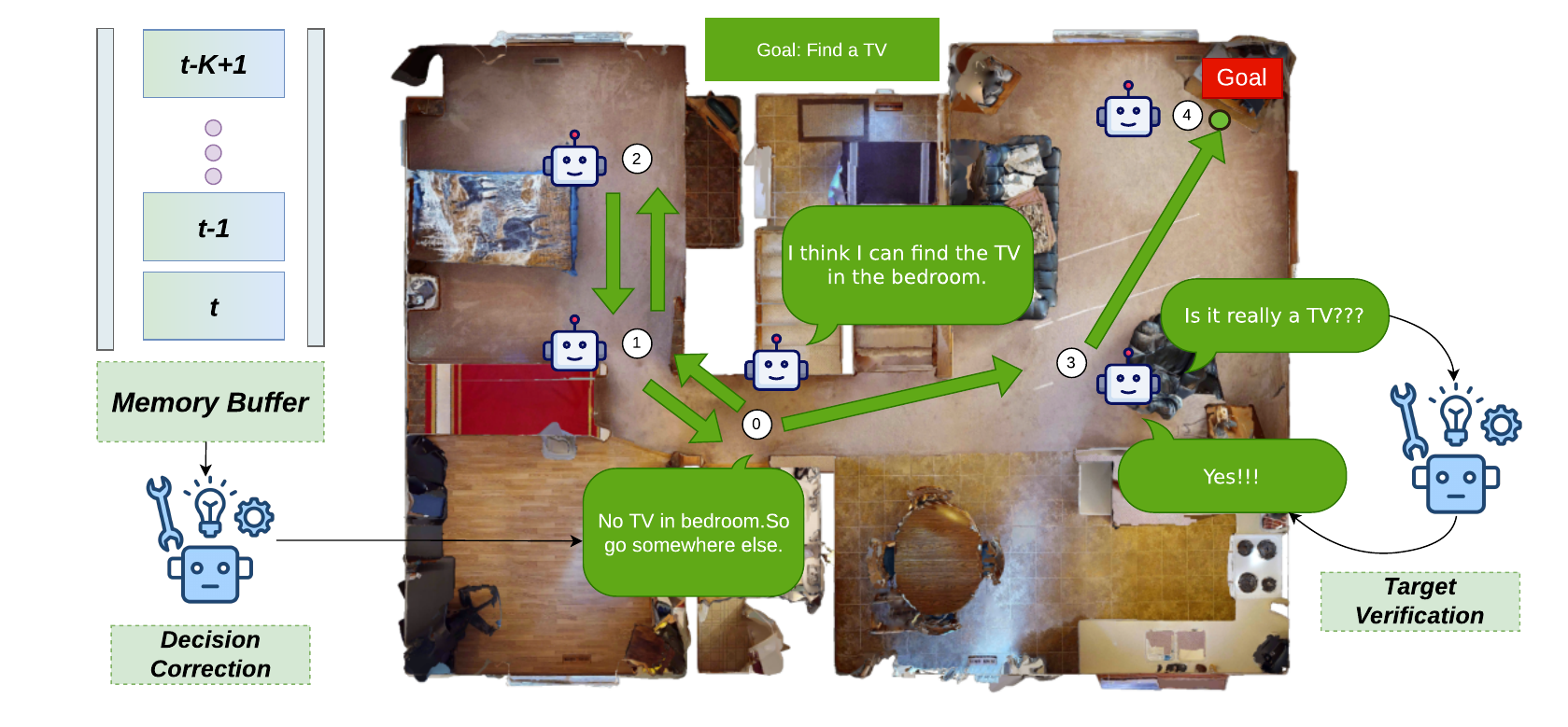}
	\caption{Illustration of ReMemNav for correcting repeated exploration and verifying potential target observations.}
	\label{fig:intro}
\end{figure}

Despite their generalization ability, existing mapless VLM-based navigators remain vulnerable to two major failure modes in complex 3D environments. First, limited first-person observations may lead to false-positive target predictions. VLMs can infer nonexistent objects from local textures or confuse visually similar objects \cite{chakraborty2025heal,dang2025ecbench}. Without an explicit verification stage, such prediction errors may cause the agent to approach an incorrect object or stop prematurely. Second, purely reactive navigation decisions often ignore historical exploration context. Since indoor object navigation is a partially observable decision-making problem that depends on temporal information \cite{chaplot2020learning}, agents that repeatedly reason only from the current observation may revisit explored regions, become trapped near dead ends, or exhibit cyclic exploration.

To address these issues, we propose ReMemNav, a training-free zero-shot object navigation framework illustrated in Fig.~\ref{fig:intro}. The framework combines lightweight semantic grounding with memory-based decision correction and target verification. RAM-derived semantic priors provide auxiliary object-level information for panoramic direction selection. When the target is not detected, ReMemNav projects recently visited states into the current navigable view and triggers memory-based correction if the selected direction overlaps with an explored region. The retrieved historical scene description is then used to revise the potentially repetitive decision. When the target is predicted to be visible, ReMemNav performs target verification on the corresponding single-view observation before executing the final approach. The resulting high-level direction is subsequently converted into collision-free movement through depth-based candidate action sampling.

Our contributions can be summarized as follows:
\begin{itemize}
	\item We propose ReMemNav, a training-free zero-shot object navigation framework that combines lightweight semantic grounding with memory-based decision correction and target verification. Semantic priors extracted by the Recognize Anything Model (RAM) provide auxiliary object-level information for panoramic navigation decisions.

	\item We develop a geometry-triggered memory-based decision correction mechanism that integrates a lightweight episodic memory buffer with projection-based historical retrieval. When the selected direction overlaps with a previously explored region, the retrieved semantic context is used to revise potentially repetitive navigation decisions.

	\item We introduce an explicit target verification mechanism that re-examines potential target observations before the final approach, reducing false-positive target decisions and premature stopping. Extensive experiments on HM3D \cite{ramakrishnan2021habitat} and MP3D \cite{chang2017matterport3d} show that ReMemNav achieves higher navigation success and exploration efficiency than representative training-free zero-shot baselines.
\end{itemize}

\begin{figure*}[t]
	\centering
	\includegraphics[width=0.95\textwidth]{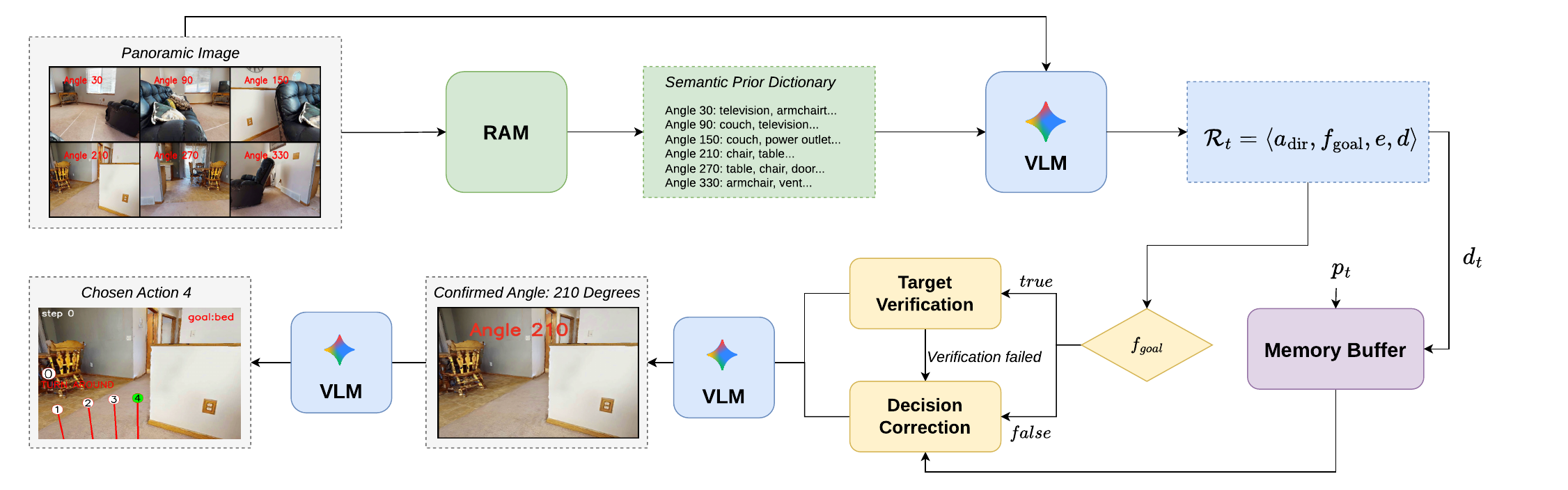}
	\caption{Overview of ReMemNav. RAM provides lightweight semantic grounding for the initial panoramic navigation decision. When the selected direction overlaps with a previously explored region, memory-based decision correction retrieves historical semantic context and revises the decision. When the target is predicted to be visible, target verification re-examines the corresponding observation before the final approach.}
	\label{fig:framework}
\end{figure*}

\section{Related Work}

\subsection{Zero-Shot and VLM-Based Object Navigation}

Traditional object navigation methods typically rely on deep
reinforcement learning
\cite{maksymets2021thda,wijmans2019dd,chen2022learning,3,4}
or imitation learning \cite{ramrakhya2022habitat} in large-scale
simulated environments. Although effective on predefined target
categories, these methods require substantial task-specific
training and often generalize poorly to unseen scenes or
open-vocabulary targets. This limitation has motivated increasing
interest in zero-shot object navigation \cite{al2022zero}. Early
approaches use pretrained vision-language models such as CLIP to
align visual observations with target descriptions and guide
exploration through image-text similarity
\cite{2,khandelwal2022simple,gadre2023cows}. However, they may
lack contextual semantics and global spatial information
\cite{sun2024prioritized}.

Map-based methods address longer-range spatial reasoning by
constructing semantic occupancy grids, visual-language maps, or
predictive maps during exploration
\cite{7,huang2023visual,zhang2024imagine}. Recent approaches
further combine these representations with large-model
commonsense reasoning for heuristic exploration in unknown
environments
\cite{6,yu2023l3mvn,yokoyama2024vlfm,kuang2024openfmnav,9,nie2025wmnav}.
Despite their effectiveness, explicit map construction and
maintenance introduce additional computational and system
complexity.

Another line of work directly employs VLMs for mapless navigation
decisions \cite{goetting2024end,5}, predicting
promising room types, directions, or actions from current visual
observations and commonsense knowledge. However, limited
first-person observations can cause perceptual ambiguity,
false-positive target predictions, and confusion between visually
similar objects. External object-relation priors have been used to
improve visual-semantic reasoning \cite{chen2023zero}. In ReMemNav, the Recognize Anything Model (RAM)
\cite{zhang2024recognize} provides lightweight directional
object tags as auxiliary semantic priors for panoramic
decision-making.

\subsection{Memory-Based Navigation and Decision Verification}

Indoor object navigation is partially observable, making historical
context important for avoiding repeated visits and cyclic
exploration. Prior studies therefore investigate temporal memory
and historical-context modeling for navigation
\cite{chen2022think,wang2025clash,gu2025doraemon}.

Re-evaluation and self-verification mechanisms have also been
introduced to revise unreliable model predictions
\cite{lin2023swiftsage,wei2025ground}. ReMemNav combines these
directions through two distinct mechanisms: geometry-triggered
retrieval of historical semantic context for correcting potentially
repetitive decisions, and independent single-view verification of
potential target observations to reduce false-positive decisions
and premature stopping.

\section{Methodology}

\subsection{Task Definition}

The zero-shot object navigation task requires an agent to explore an unknown indoor environment and navigate to any target instance belonging to a given category $c$. At each time step $t$, the agent receives an RGB-D observation $O_t$ and its current pose $P_t$. The observation consists of multi-view RGB images and their corresponding depth maps, while the pose provides the agent position and orientation required for geometric action execution and historical-node projection.

Based on the current observation and navigation history, the agent predicts an action $a_t=(r_t,\theta_t)$, where $r_t$ denotes the movement distance and $\theta_t$ denotes the relative yaw angle. An episode is considered successful when the agent actively issues the stop action within the benchmark-defined distance threshold $d_{\text{thres}}$ from a valid target instance. The ground-truth target position used for evaluation is not provided to the navigation agent.

\subsection{System Overview}

The overall architecture of ReMemNav is illustrated in Fig.~\ref{fig:framework}. At each time step, the agent acquires six RGB-D observations corresponding to the relative yaw angles $\mathcal{A}=\{30^\circ,90^\circ,150^\circ,210^\circ,270^\circ,330^\circ\}$. RAM first extracts directional object-level semantic priors from the six RGB views. The VLM then integrates the panoramic observation, semantic priors, and target category to generate an initial exploration direction, a target-presence flag, a decision rationale, and a panoramic scene description.

ReMemNav subsequently follows two state-conditioned decision paths. When the target is not predicted to be visible, the memory-based decision correction mechanism geometrically checks whether the selected direction may revisit a previously explored region and retrieves the corresponding historical context for decision revision. When the target is predicted to be visible, the target verification mechanism re-examines the selected single-view observation before the final approach. Finally, the resulting high-level direction is mapped to a collision-free low-level movement using depth-based candidate action sampling.


\subsection{Lightweight Semantic Grounding with RAM}

Limited first-person observations can make VLM-based navigation vulnerable to perceptual ambiguity and confusion between visually similar objects. We therefore adopt the pretrained Recognize Anything Model (RAM) as a lightweight semantic grounding module rather than treating it as an independent navigation policy.

At time step $t$, the six RGB observations are denoted as $\mathcal{V}_t=\{V_t^i\}_{i=1}^{6}$, where $\theta_i\in\mathcal{A}$. RAM extracts a set of high-confidence object labels from each view. These labels are associated with their corresponding viewing directions to form a directional semantic prior $\mathcal{S}_t=\langle(\theta_i,\operatorname{RAM}(V_t^i))\rangle_{i=1}^{6}$. For example, an element in $\mathcal{S}_t$ may be represented as
\textit{Angle $30^\circ$: armchair, lamp, carpet, couch}.
These structured labels provide auxiliary object-level information for panoramic spatial reasoning.

The six RGB views are further concatenated into a panoramic observation $V_t^{\mathrm{pan}}$. Given the target category $c$ and navigation prompt $q_{\mathrm{nav}}$, the VLM produces a structured response:
\begin{equation}
    \mathcal{R}_t
    =
    \operatorname{VLM}
    \left(
        V_t^{\mathrm{pan}},
        \mathcal{S}_t,
        c,
        q_{\mathrm{nav}}
    \right)
    =
    \left\langle
        a_t^{\mathrm{dir}},
        f_t^{\mathrm{goal}},
        e_t,
        d_t
    \right\rangle ,
    \label{eq:initial_decision}
\end{equation}
where $a_t^{\mathrm{dir}} \in \mathcal{A}$ is the selected navigation direction, $f_t^{\mathrm{goal}} \in \{0,1\}$ indicates whether the target is predicted to be visible, $e_t$ is the decision rationale, and $d_t$ is the panoramic description of the current scene.

\subsection{Memory-Based Decision Correction}

Purely reactive VLM navigators make each decision primarily from the current observation and may repeatedly select previously explored regions. ReMemNav addresses this problem through a memory-based decision correction mechanism consisting of a bounded episodic memory, geometry-triggered historical retrieval, and history-conditioned decision revision.

\paragraph{Episodic Memory Construction.}

At each decision step, ReMemNav stores the current world-coordinate position
$\mathbf{p}_t^{w}=[x_t,y_t,z_t]^\top$
and the VLM-generated panoramic scene description $d_t$. The episodic memory is represented as an ordered sequence:
\begin{equation}
    \mathcal{M}_t
    =
    \left\langle
        \left(
            \mathbf{p}_i^{w},
            d_i
        \right)
    \right\rangle_
    {i=\max(1,t-K+1)}^{t},
    \label{eq:memory}
\end{equation}
where $\mathbf{p}_i^{w}$ denotes the three-dimensional position of the $i$-th visited node in the world coordinate system, $d_i$ is its corresponding scene description, and $K$ is the maximum memory capacity.

The memory is updated using a first-in-first-out strategy:
\begin{equation}
    \mathcal{M}_t
    =
    \operatorname{Tail}_{K}
    \left(
        \mathcal{M}_{t-1}
        \oplus
        \left(
            \mathbf{p}_t^{w},
            d_t
        \right)
    \right),
    \label{eq:memory_update}
\end{equation}
where $\oplus$ denotes appending a new node to the end of the ordered sequence and $\operatorname{Tail}_{K}(\cdot)$ retains only the most recent $K$ nodes. We empirically set $K=10$ in the main experiments.

Unlike an explicit occupancy or semantic map, $\mathcal{M}_t$ stores only recently visited positions and their corresponding language descriptions. It neither represents the complete environment nor provides information about unexplored regions.

\paragraph{Geometry-Triggered Historical Retrieval.}

When $f_t^{\mathrm{goal}}=0$, ReMemNav checks whether the initially selected direction $a_t^{\mathrm{dir}}$ may lead the agent toward a previously explored region. To avoid matching the current node with itself, the retrieval process uses the historical memory $\mathcal{M}_{t-1}$.

Each historical node $\mathbf{p}_i^{w}\in\mathcal{M}_{t-1}$ is projected into the current camera view:
\begin{equation}
    \lambda_i
    \begin{bmatrix}
        u_i \\
        v_i \\
        1
    \end{bmatrix}
    =
    \mathbf{K}_{\mathrm{cam}}
    \mathbf{T}_{w\rightarrow t}
    \begin{bmatrix}
        \mathbf{p}_i^{w} \\
        1
    \end{bmatrix},
    \label{eq:projection}
\end{equation}
where $\mathbf{K}_{\mathrm{cam}}$ is the camera intrinsic matrix, $\mathbf{T}_{w\rightarrow t}$ transforms a world-coordinate point into the current camera coordinate system, $(u_i,v_i)$ is the projected pixel coordinate, and $\lambda_i$ is the corresponding projected depth.

The set of historical nodes overlapping with the selected navigable direction is defined as
\begin{equation}
    \mathcal{I}_t^{\mathrm{hit}}
    =
    \left\{
        i \,\middle|\,
        \substack{
            \lambda_i>0,\ (u_i,v_i)\in
            \operatorname{Mask}_{\mathrm{nav}}(a_t^{\mathrm{dir}}) \\
            |\Delta\theta_i|<\tau_{\theta}
        }
    \right\},
    \label{eq:trigger_set}
\end{equation}
where $\operatorname{Mask}_{\mathrm{nav}}(a_t^{\mathrm{dir}})$ denotes the collision-free image region associated with the selected direction, $\Delta\theta_i$ is the angular deviation between the historical node and $a_t^{\mathrm{dir}}$, and $\tau_{\theta}$ is the angular threshold. We set $\tau_\theta=40^\circ$, approximately half of the camera's $79^\circ$ horizontal field of view, so that a historical node is considered relevant when it falls within the visual range associated with the selected direction.

Memory-based correction is triggered only when
$\mathcal{I}_t^{\mathrm{hit}}\neq\varnothing$.
When multiple historical nodes satisfy the trigger condition, ReMemNav retrieves the furthest matched node:
\begin{equation}
    i^{*}
    =
    \mathop{\arg\max}_{i\in\mathcal{I}_t^{\mathrm{hit}}}
    \left\|
        \mathbf{p}_i^{w}
        -
        \mathbf{p}_t^{w}
    \right\|_2 .
    \label{eq:retrieval}
\end{equation}
This deterministic retrieval rule selects the deepest previously visited location along the intended direction, providing historical evidence that the corresponding region has already been explored.

\paragraph{History-Conditioned Decision Revision.}

The retrieved historical description $d_{i^{*}}$ is supplied to the VLM through a correction prompt $q_{\mathrm{corr}}$. The corrected exploration direction is defined as
\begin{equation}
    \widetilde{a}_t^{\mathrm{dir}}
    =
    \begin{cases}
        \operatorname{VLM}_{\mathrm{dir}}
        \left(
            V_t^{\mathrm{pan}},
            \mathcal{S}_t,
            c,
            d_{i^{*}},
            q_{\mathrm{corr}}
        \right),
        &
        \mathcal{I}_t^{\mathrm{hit}}
        \neq
        \varnothing,
        \\[4pt]
        a_t^{\mathrm{dir}},
        &
        \mathcal{I}_t^{\mathrm{hit}}
        =
        \varnothing,
    \end{cases}
    \label{eq:corrected_decision}
\end{equation}
where $\operatorname{VLM}_{\mathrm{dir}}(\cdot)$ denotes the navigation-direction field returned by the VLM. If no historical node satisfies the trigger condition, the original direction is retained.

\begin{figure*}[t]
    \centering
    \includegraphics[width=0.95\textwidth]{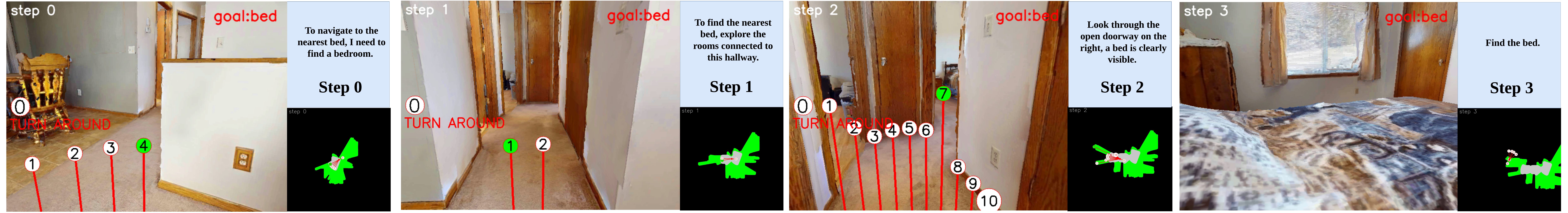}
    \caption{VLM-guided low-level action selection. Collision-free candidate action rays are generated from the depth observation and overlaid on the RGB image, after which the VLM selects the action that best follows the finalized high-level direction.}
    \label{fig:process}
\end{figure*}

\subsection{Target Verification for False-Positive Rejection}

When $f_t^{\mathrm{goal}}=1$, the initial VLM response indicates that the target may be visible in direction $a_t^{\mathrm{dir}}$. However, this prediction may still be caused by visual similarity, local texture patterns, or ambiguous observations. ReMemNav therefore performs an explicit verification step before entering the final target-approach stage.

Let $j(a_t^{\mathrm{dir}})$ denote the view index corresponding to the selected direction. Given the selected single-view RGB observation and a verification prompt $q_{\mathrm{ver}}$, the verification result is defined as
\begin{equation}
    v_t
    =
    \operatorname{VLM}
    \left(
        V_t^{j(a_t^{\mathrm{dir}})},
        c,
        q_{\mathrm{ver}}
    \right)
    \in
    \{0,1\},
    \label{eq:verification}
\end{equation}
where $v_t=1$ indicates that the target prediction is verified and $v_t=0$ indicates that it is rejected.

The verified target flag is computed as
\begin{equation}
    \widetilde{f}_t^{\mathrm{goal}}
    =
    f_t^{\mathrm{goal}}
    \land
    v_t .
    \label{eq:verified_goal}
\end{equation}
Only when $\widetilde{f}_t^{\mathrm{goal}}=1$ does the agent enter the final target-approach stage. When target verification fails, the original direction
$a_t^{\mathrm{dir}}$ is further processed by the memory-based
decision correction mechanism in Eq.~(7).

The target verification process does not access the ground-truth target coordinates. The benchmark evaluator independently determines whether the stop action satisfies the predefined success-distance criterion.

\subsection{Low-Level Action Execution}

As illustrated in Fig.~\ref{fig:process}, after obtaining the final high-level direction, the agent converts it into a collision-free physical movement. Let $\bar{a}_t^{\mathrm{dir}}$ denote the finalized direction:
\begin{equation}
\bar{a}^{\mathrm{dir}}_t =
\begin{cases}
\tilde{a}^{\mathrm{dir}}_t,
& f_t^{\mathrm{goal}} = 0
  \ \lor\
  \tilde{f}_t^{\mathrm{goal}} = 0,\\
a_t^{\mathrm{dir}},
& \tilde{f}_t^{\mathrm{goal}} = 1.
\end{cases}
\end{equation}

Following the depth-based visual action mapping strategy used in previous VLM-driven navigation methods \cite{goetting2024end,nie2025wmnav}, ReMemNav constructs a traversability mask from the depth observation associated with $\bar{a}_t^{\mathrm{dir}}$. A set of candidate action rays is sampled from the robot base toward collision-free locations:
\begin{equation}
    \begin{aligned}
        \mathcal{C}_t
        &=
        \left\{
            (r_{t,m},\theta_{t,m})
        \right\}_{m=1}^{M} \\
        &=
        \operatorname{SampleRays}
        \left(
            D_t^{j(\bar{a}_t^{\mathrm{dir}})},
            \operatorname{Mask}_{\mathrm{nav}}
        \right),
    \end{aligned}
    \label{eq:candidate_actions}
\end{equation}
where $D_t^{j(\bar{a}_t^{\mathrm{dir}})}$ is the depth observation corresponding to the selected direction and $M$ is the number of sampled candidate rays.

The candidate trajectories are projected onto the RGB observation and indexed with numerical labels. The VLM then selects one candidate according to the current visual observation and navigation objective:
\begin{equation}
    \begin{aligned}
        m_t^{*}
        &=
        \operatorname{VLMSelect}
        \left(
            V_t^{j(\bar{a}_t^{\mathrm{dir}})},
            \mathcal{C}_t,
            c
        \right), \\
        a_t
        &=
        \left(
            r_{t,m_t^{*}},
            \theta_{t,m_t^{*}}
        \right).
    \end{aligned}
    \label{eq:action_selection}
\end{equation}
The selected polar-coordinate action is subsequently mapped to the simulator motion interface for execution.

\paragraph{Goal-Position Estimation and Stopping.}
When the target prediction is successfully verified, the agent
does not immediately issue a stop action. Instead, the VLM
selects the candidate navigation arrow that leads toward the
verified target. Let $(r_{t,m_t^*}, \theta_{t,m_t^*})$ denote
the selected polar-coordinate action. The endpoint of the
selected arrow is transformed into the world coordinate system
and used as an estimated goal position:
\begin{equation}
\mathbf{g}_t^w
=
\mathbf{p}_t^w
+
\mathbf{R}_t^w
\begin{bmatrix}
r_{t,m_t^*}\cos\theta_{t,m_t^*} \\
r_{t,m_t^*}\sin\theta_{t,m_t^*} \\
0
\end{bmatrix},
\end{equation}
where $\mathbf{p}_t^w$ and $\mathbf{R}_t^w$ denote the current
agent position and orientation in the world coordinate system,
respectively.

The estimated goal positions obtained during target approach
are stored in a goal-position buffer $\mathcal{G}_t$. Their
average is computed as
\begin{equation}
\bar{\mathbf{g}}_t^w
=
\frac{1}{|\mathcal{G}_t|}
\sum_{\mathbf{g}_i^w \in \mathcal{G}_t}
\mathbf{g}_i^w.
\end{equation}
At each subsequent step, the agent computes the planar distance
between its current position and the averaged estimated goal
position:
\begin{equation}
d_t^{\mathrm{goal}}
=
\left\|
\left(
\mathbf{p}_t^w-\bar{\mathbf{g}}_t^w
\right)_{xy}
\right\|_2.
\end{equation}
The agent continues approaching the estimated goal position
until $d_t^{\mathrm{goal}} < d_{\mathrm{stop}}$, where
$d_{\mathrm{stop}}=0.8\,\mathrm{m}$. Once this condition is
satisfied, the agent executes \texttt{PolarAction.stop};
otherwise, it continues the target-approach process. Note that $d_{\mathrm{stop}}$ is the internal
stopping threshold used by the agent, whereas
$d_{\mathrm{thres}}=1.0\,\mathrm{m}$ is the benchmark-defined
success threshold used only by the evaluator. We adopt the
more conservative internal threshold to provide a margin for
goal-position estimation and action-execution errors.

\begin{table*}[t]
	\centering
	\small
	\begin{tabular}{lcccccccc}
		\toprule
		\multirow{2}{*}{Methods}                 & \multirow{2}{*}{TF} & \multirow{2}{*}{ZS} & \multicolumn{2}{c}{MP3D} & \multicolumn{2}{c}{HM3D\_v0.1} & \multicolumn{2}{c}{HM3D\_v0.2}                                                 \\
		\cmidrule(lr){4-5} \cmidrule(lr){6-7} \cmidrule(lr){8-9}
		                                         &                     &                     & SR                       & SPL                            & SR                             & SPL           & SR            & SPL           \\
		\midrule
		{Habitat-Web}~\cite{ramrakhya2022habitat} & \ding{55}           & \ding{55}           & 31.6                     & 8.5                            & 41.5                           & 16.0          & --            & --            \\
		{OVRL}~\cite{3}                           & \ding{55}           & \ding{55}           & 28.6                     & 7.4                            & --                             & --            & --            & --            \\
		{ZSON}~\cite{2}                           & \ding{55}           & \pmb{$\checkmark$}  & 15.3                     & 4.8                            & 25.5                           & 12.6          & --            & --            \\
		{PixNav}~\cite{5}                         & \ding{55}           & \pmb{$\checkmark$}  & --                       & --                             & 37.9                           & 20.5          & --            & --            \\
		{PSL}~\cite{sun2024prioritized}           & \ding{55}           & \pmb{$\checkmark$}  & 18.9                     & 6.4                            & 42.4                           & 19.2          & --            & --            \\
		{SGM}~\cite{zhang2024imagine}             & \ding{55}           & \pmb{$\checkmark$}  & 37.7                     & 14.7                           & 60.2                           & 30.8          & --            & --            \\
		{VLFM}~\cite{yokoyama2024vlfm}            & \ding{55}           & \pmb{$\checkmark$}  & 36.4                     & 17.5                           & 52.5                           & 30.4          & 62.6          & 31.0          \\
		{Chain-of-Search}~\cite{chen2026chain}    & \ding{55}           & \pmb{$\checkmark$}  & 37.6                     & 17.6                           & 55.9                           & 29.1          & --            & --            \\
		\midrule
		{CoW}~\cite{gadre2023cows}                & \pmb{$\checkmark$}  & \pmb{$\checkmark$}  & 9.2                      & 4.9                            & --                             & --            & --            & --            \\
		{ESC}~\cite{6}                            & \pmb{$\checkmark$}  & \pmb{$\checkmark$}  & 28.7                     & 14.2                           & 39.2                           & 22.3          & --            & --            \\
		{L3MVN}~\cite{yu2023l3mvn}                & \pmb{$\checkmark$}  & \pmb{$\checkmark$}  & --                       & --                             & 50.4                           & 23.1          & 36.3          & 15.7          \\
		{TopV-Nav}~\cite{9}                       & \pmb{$\checkmark$}  & \pmb{$\checkmark$}  & 35.2                     & 16.4                           & 53.0                           & 29.8          & --            & --            \\
		{SG-Nav}~\cite{yin2024sg}                 & \pmb{$\checkmark$}  & \pmb{$\checkmark$}  & 40.2                     & 16.0                           & 54.0                           & 24.9          & 49.6          & 25.5          \\
		{VLMNav}~\cite{goetting2024end}           & \pmb{$\checkmark$}  & \pmb{$\checkmark$}  & --                       & --                             & 50.4                           & 21.0          & --            & --            \\
		{InstructNav}~\cite{long2024instructnav}  & \pmb{$\checkmark$}  & \pmb{$\checkmark$}  & --                       & --                             & 58.0                           & 20.9          & --            & --            \\
		{UniGoal}~\cite{yin2025unigoal}           & \pmb{$\checkmark$}  & \pmb{$\checkmark$}  & 41.0                     & 16.4                           & 54.5                           & 25.1          & --            & --            \\
		{MFNP}~\cite{zhang2025multi}              & \pmb{$\checkmark$}  & \pmb{$\checkmark$}  & 41.1                     & 15.4                           & 58.3                           & 26.7          & --            & --            \\
		{PanoNav}~\cite{jin2026panonav}           & \pmb{$\checkmark$}  & \pmb{$\checkmark$}  & --                       & --                             & 43.5                           & 23.7          & --            & --            \\
		{EIK-Nav}~\cite{jiang2026eiknav}          & \pmb{$\checkmark$}  & \pmb{$\checkmark$}  & 37.1                     & 17.4                           & 54.8                           & 31.5          & --            & --            \\
		\midrule
		\textbf{ReMemNav (Ours)}                 & \pmb{$\checkmark$}  & \pmb{$\checkmark$}  & \textbf{49.8}            & \textbf{24.3}                  & \textbf{60.0}                  & \textbf{36.8} & \textbf{67.8} & \textbf{36.6} \\
		\bottomrule
	\end{tabular}
	\caption{Comparison with representative object navigation methods on HM3D and MP3D. TF refers to training-free, and ZS refers to zero-shot.}
	\label{tab:main_results_detailed}
\end{table*}

\begin{figure*}[t]
    \centering
    \includegraphics[width=0.95\textwidth]{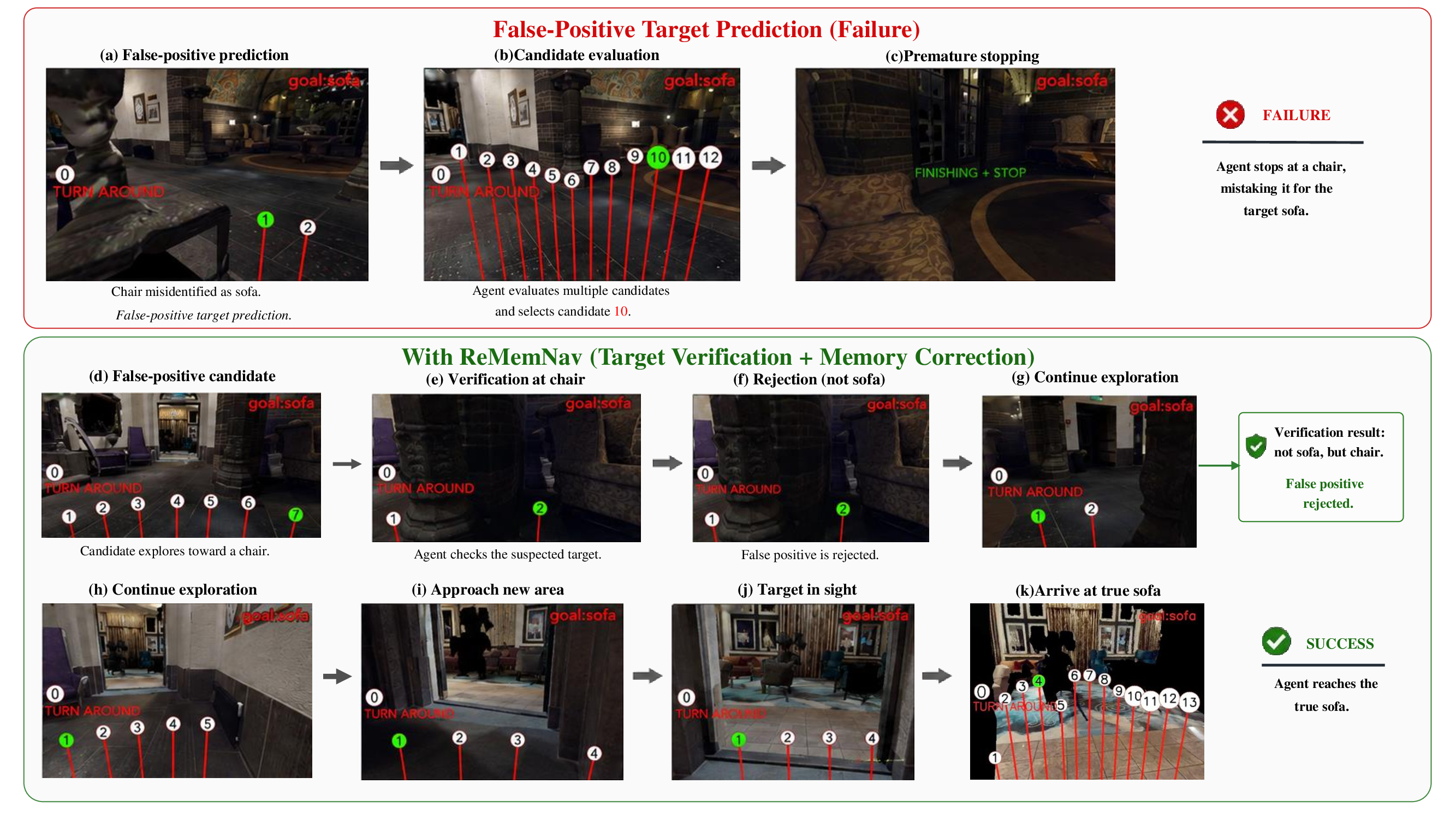}
    \caption{Qualitative comparison of navigation behaviors on the
    same episode. In the top row, the ablated agent without target
    verification incorrectly
    identifies a chair as the target sofa. It consequently approaches
    the false-positive target and stops prematurely. In contrast, full ReMemNav re-examines the
    suspected target and
    correctly rejects the chair as a false positive. The agent then
    continues exploring the environment and eventually reaches a valid
    sofa instance, completing the episode successfully.}
    \label{fig:experimental_results}
\end{figure*}

\begin{table}[t]
    \centering
    \small
    \begin{tabular}{lcc}
        \toprule
        Method
        & SR (\%) $\uparrow$
        & SPL (\%) $\uparrow$ \\
        \midrule
        VLMNav
        & 41.5
        & 11.7 \\
        UniGoal
        & 43.3
        & 21.7 \\
        InstructNav
        & 53.5
        & 21.1 \\
        \midrule
        \textbf{ReMemNav (Ours)}
        & \textbf{56.2}
        & \textbf{21.8} \\
        \bottomrule
    \end{tabular}
    \caption{Comparison under a common VLM backbone.
    All methods are evaluated with Qwen3-VL-4B on HM3D v0.2 under the same benchmark configuration.}
    \label{tab:fair_comparison}
\end{table}

\begin{table}[t]
    \centering
    \small
    \begin{tabular}{lcc}
        \toprule
        Variant
        & SR (\%) $\uparrow$
        & SPL (\%) $\uparrow$ \\
        \midrule
        \textbf{Full ReMemNav}
        & \textbf{56.2}
        & \textbf{21.8} \\

        w/o Semantic Grounding
        & 55.4
        & 21.3 \\

        w/o Memory-Based Correction
        & 52.8
        & 21.4 \\

        w/o Target Verification
        & 45.9
        & 21.0 \\
        \bottomrule
    \end{tabular}
    \caption{Module-wise ablation on HM3D v0.2.
    Semantic grounding is implemented using RAM. Memory-based correction jointly includes the bounded episodic memory and history-conditioned decision revision.}
    \label{tab:module_ablation}
\end{table}


\section{Experiments}

\subsection{Datasets and Evaluation Metrics}




\paragraph{Datasets.}
We evaluate ReMemNav on HM3D v0.1, HM3D v0.2, and
Matterport3D (MP3D). HM3D v0.1
\cite{ramakrishnan2021habitat} contains 20 validation
environments and 2,000 episodes across six target categories,
while HM3D v0.2 contains 1,000 episodes with improved semantic
annotations and geometry. MP3D \cite{chang2017matterport3d}
contains 11 validation scenes and 2,195 episodes across 21
target categories.

\paragraph{Evaluation Metrics.}
We evaluate navigation performance using Success Rate (SR)
and Success weighted by Path Length (SPL). SR is defined as
$SR=\frac{1}{N}\sum_{i=1}^{N}S_i$, where $N$ is the number
of episodes and $S_i \in \{0,1\}$
indicates success in episode $i$.

SPL jointly measures success and path efficiency and is defined as
$SPL=\frac{1}{N}\sum_{i=1}^{N}S_i\frac{L_i}{\max(P_i,L_i)}$,
where $L_i$ and $P_i$ denote the oracle geodesic distance and
executed path length, respectively. $L_i$ is computed by the
benchmark evaluator and is not available to the agent.

\subsection{Implementation Details}

Each navigation episode is limited to a maximum of 40 decision steps. The agent has a cylindrical body with a radius of 0.18\,m and a height of 0.88\,m. It is equipped with an egocentric RGB-D camera with a resolution of $640 \times 480$, a horizontal field of view of $79^\circ$, and a downward tilt of 0.25 radians, approximately $14^\circ$, to capture ground-level depth information for traversability estimation.

The capacity of the bounded episodic memory is set to $K=10$, unless otherwise specified. Following the benchmark protocol, an episode is considered successful only when the agent actively executes the stop action within $d_{\text{thres}}=1.0$\,m of a valid target instance.

The main benchmark results are obtained using Gemini-3-Flash as the VLM backbone. Module-wise ablations, common-backbone comparisons, and latency measurements are conducted using Qwen3-VL-4B. All VLMs are used without navigation-specific training or fine-tuning. Additional results on backbone sensitivity, memory capacity, and robustness to pose noise and odometry drift, together with prompts and implementation details, are provided in the supplementary material.

\subsection{Comparison with Existing Methods}

We compare ReMemNav with representative learning-based and training-free zero-shot object navigation methods on MP3D, HM3D v0.1, and HM3D v0.2. The compared approaches include feature-alignment methods, map-based navigation systems, and recent VLM-driven mapless navigators.


As shown in Table~\ref{tab:main_results_detailed}, ReMemNav achieves the best reported SR and SPL among the compared training-free zero-shot methods on MP3D and HM3D v0.2. Compared with the strongest reported training-free baselines on the corresponding benchmarks, ReMemNav improves SR and SPL by 8.7 and 6.9 percentage points on MP3D, respectively, and by 18.2 and 11.1 percentage points on HM3D v0.2.

When compared with all listed methods, including approaches that require task-specific training, ReMemNav obtains the highest SR and SPL on MP3D and HM3D v0.2. On HM3D v0.1, ReMemNav achieves a competitive SR of 60.0\% and the highest SPL of 36.8\% among the compared methods. These results demonstrate the strong navigation performance of the proposed framework across different simulated indoor benchmarks.

\subsection{Comparison under a Common VLM Backbone}

Different navigation systems may employ VLMs with substantially different reasoning capacities. To reduce this confounding factor, we reimplement VLMNav \cite{goetting2024end}, UniGoal \cite{yin2025unigoal}, InstructNav \cite{long2024instructnav}, and ReMemNav using the same Qwen3-VL-4B backbone on HM3D v0.2. All methods follow the same episode set, success criterion, and evaluation protocol.

As shown in Table~\ref{tab:fair_comparison}, ReMemNav achieves the highest SR of 56.2\% and SPL of 21.8\% under the common VLM backbone. Compared with InstructNav, the strongest baseline in SR, ReMemNav improves SR by 2.7 percentage points and SPL by 0.7 percentage points. These results indicate that the performance gains cannot be attributed solely to the use of Gemini-3-Flash and that the proposed semantic grounding, memory-based correction, and target verification mechanisms remain effective with an open-source VLM backbone. Results with additional VLM backbones are reported in the supplementary material.

\subsection{Module-Wise Ablation}

To examine the contribution of each functional component, we independently remove semantic grounding, memory-based decision correction, and target verification from the complete ReMemNav framework. All variants are evaluated on HM3D v0.2 using Qwen3-VL-4B under the same experimental configuration.

As shown in Table~\ref{tab:module_ablation}, the complete framework achieves the best overall performance. Removing RAM-based semantic grounding decreases SR from 56.2\% to 55.4\% and SPL from 21.8\% to 21.3\%, indicating that directional semantic priors provide modest but consistent auxiliary grounding.

Disabling memory-based decision correction reduces SR by 3.4 percentage points. Since the episodic memory buffer supplies the historical context required for decision revision, the memory buffer and its history-conditioned correction process are treated as one coupled functional component in this ablation.

The largest degradation occurs when target verification is removed: SR decreases from 56.2\% to 45.9\%, while SPL decreases from 21.8\% to 21.0\%. This result is consistent with the role of explicit verification in rejecting false-positive target predictions and reducing premature stopping. A qualitative comparison of these navigation behaviors is shown in Fig.~\ref{fig:experimental_results}. Overall, the three components provide complementary benefits, with target verification contributing most strongly to navigation success.

\subsection{System Overhead and Latency Analysis}

We measure the average latency of one complete navigation decision cycle on HM3D v0.2 using a single NVIDIA RTX 4090 GPU and the Qwen3-VL-4B backbone. The same hardware and backbone configuration is used for all reproduced methods.

ReMemNav requires an average of 5660\,ms per decision cycle. This latency is comparable to VLMNav (5957\,ms) and lower than those of the reproduced UniGoal (8207\,ms) and InstructNav (14031\,ms) implementations.

Within ReMemNav, RAM-based semantic grounding requires approximately 453\,ms, while geometric memory retrieval requires approximately 0.6\,ms. When triggered, memory-based decision correction and target verification introduce approximately 3145\,ms and 2839\,ms of additional latency, respectively, because each mechanism requires an additional VLM inference. These mechanisms are conditionally activated only at decision steps that satisfy their corresponding trigger conditions. The reported measurements characterize per-decision computational cost rather than total episode wall-clock time.

\section{Conclusion}

In this work, we presented ReMemNav, a training-free framework for zero-shot object navigation that combines lightweight semantic grounding, memory-based decision correction, and target verification. RAM-derived semantic priors provide auxiliary object-level information for panoramic navigation decisions, while geometry-triggered historical retrieval reduces repeated
exploration and explicit verification suppresses false-positive
target predictions and premature stopping. Experiments on HM3D and MP3D demonstrate strong navigation performance and improved exploration efficiency. The current evaluation is conducted in static indoor environments. Future work will explore more scalable memory representations for long-horizon navigation, extend the framework to dynamic and more diverse environments, and validate its effectiveness on real robotic systems.
\bibliography{references}

\end{document}